\documentclass[10pt,twocolumn,letterpaper]{article}

\usepackage{iccv}
\usepackage{times}
\usepackage{epsfig}
\usepackage{graphicx}
\usepackage{amsmath}
\usepackage{amssymb}

\usepackage{algorithm}
\usepackage{algpseudocode}% http://ctan.org/pkg/algorithmicx
\algrenewcommand\algorithmicindent{.9em}%
\usepackage{enumerate}
\usepackage{multirow}
\usepackage{enumitem}
\usepackage{makecell}
\usepackage{tabulary}
\usepackage{pifont}
\usepackage{boldline}

\usepackage[breaklinks=true,bookmarks=false]{hyperref}

\iccvfinalcopy % *** Uncomment this line for the final submission

\def\iccvPaperID{****} % *** Enter the ICCV Paper ID here

\ificcvfinal\fi
\begin{document}

%%%%%%%%% TITLE
\title{Relation Distillation Networks for Video Object Detection\thanks{{\small This work was performed at JD AI Research.}}}

\author{Jiajun Deng$^\dagger$, Yingwei Pan$^\ddagger$, Ting Yao$^\ddagger$, Wengang Zhou$^\dagger$, Houqiang Li$^\dagger$, and Tao Mei$^\ddagger$\\
\small $^\dagger$ CAS Key Laboratory of GIPAS, University of Science and Technology of China, Hefei, China\\
\small $^\ddagger$JD AI Research, Beijing, China
\\
{\tt\scriptsize \{djiajun1206, panyw.ustc, tingyao.ustc\}@gmail.com, \{zhwg, lihq\}@ustc.edu.cn, tmei@jd.com}
}

\maketitle
\thispagestyle{empty}

\begin{abstract}

  It has been well recognized that modeling object-to-object relations would be helpful for object detection. Nevertheless, the problem is not trivial especially when exploring the interactions between objects to boost video object detectors. The difficulty originates from the aspect that reliable object relations in a video should depend on not only the objects in the present frame but also all the supportive objects extracted over a long range span of the video. In this paper, we introduce a new design to capture the interactions across the objects in spatio-temporal context. Specifically, we present Relation Distillation Networks (RDN) --- a new architecture that novelly aggregates and propagates object relation to augment object features for detection. Technically, object proposals are first generated via Region Proposal Networks (RPN). RDN then, on one hand, models object relation via multi-stage reasoning, and on the other, progressively distills relation through refining supportive object proposals with high objectness scores in a cascaded manner. The learnt relation verifies the efficacy on both improving object detection in each frame and box linking across frames. Extensive experiments are conducted on ImageNet VID dataset, and superior results are reported when comparing to state-of-the-art methods. More remarkably, our RDN achieves 81.8\% and 83.2\% mAP with ResNet-101 and ResNeXt-101, respectively. When further equipped with linking and rescoring, we obtain to-date the best reported mAP of 83.8\% and 84.7\%.

\end{abstract}

\section{Introduction}
%%%%%%%%% BODY TEXT\section{Introduction}
The advances in Convolutional Neural Networks (CNN) have successfully pushed the limits and improved the state-of-the-art technologies of image and video understanding \cite{he2016resnet,hu2018squeeze,huang2017densely,krizhevsky2012imagenet,li2018recurrent,li2018unified,pan2016jointly,pan2016learning,qiu2017learning,simonyan2014two,Simonyan:ICLR15,Szegedy:CVPR15}. As one of the most fundamental tasks, object detection in still images has attracted a surge of research interests and the recent methods \cite{Cai_2018_CVPR,dai2016r,girshick2015fast,he2017mask,ren2015faster} mostly proceed along the region-based detection paradigm which is derived from the work of R-CNN \cite{girshick2014rich}. In a further step to localize and recognize objects in videos, video object detection explores spatio-temporal coherence to boost detectors generally through two directions of box-level association \cite{feichtenhofer2017detect,han2016seq,kang2017object,kang2017t} and feature aggregation \cite{wang2018manet,xiao2018matchtrans,zhu2018towards,zhu2017fgfa}. The former delves into the association across bounding boxes from consecutive frames to generate tubelets. The latter improves per-frame features by aggregation of nearby features. Regardless of these different recipes for enhancing video object detection, a common issue not fully studied is the exploitation of object relation, which is well believed to be helpful for detection.

\begin{figure}[!tb]
\centering {\includegraphics[width=0.43\textwidth]{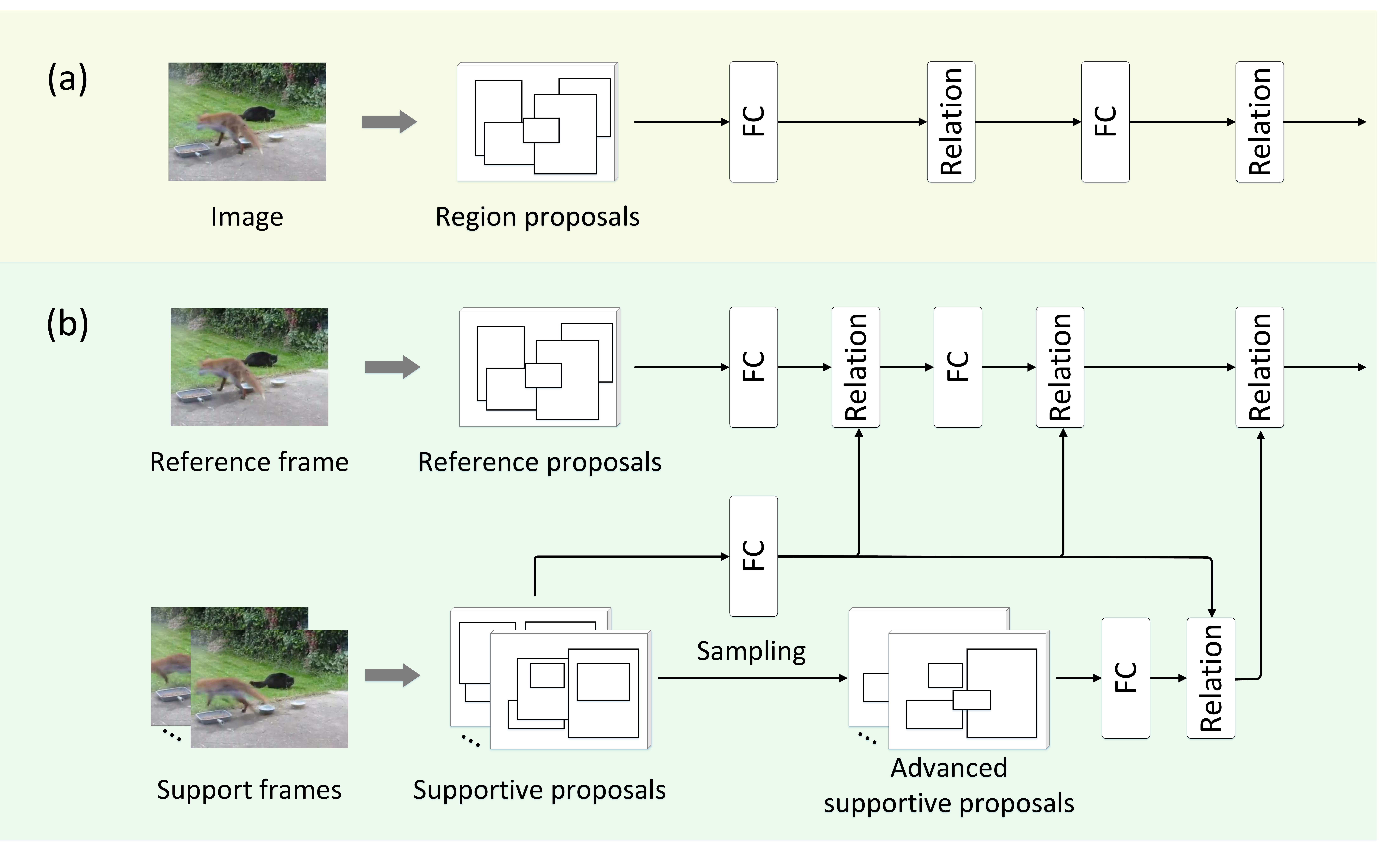}}
\vspace{-0.05in}
\caption{\small Modeling object relations by employing (a) stacked relation within an image and (b) distillation in an cascade manner across video frames.}
\label{fig:fig1}
\vspace{-0.25in}
\end{figure}

Object relations characterize the interactions or geometric positions between objects. In the literature, there has been strong evidences on the use of object relation to support various vision tasks, e.g., recognition \cite{wang2018videos}, object detection \cite{hu2018relation}, cross-domain detection \cite{cai2019exploring}, and image captioning \cite{yao2018exploring}. One representative work that employs object relation is \cite{hu2018relation} for object detection in images. The basic idea is to measure relation features of one object as the weighted sum of appearance features from other objects in the image and the weights reflect object dependency in terms of both appearance and geometry information. A stacked relation module as shown in Figure \ref{fig:fig1}(a) aggregates relation features and augments the object features in a multi-step fashion. The method verifies the merit on modeling object relation to eventually enhance image object detection. Nevertheless, the extension of mining object relation in an image to in a video is very challenging due to the complex spatio-temporal context. Both the objects in the reference frame and all the supportive objects extracted from nearby frames should be taken into account. This distinction leads to a huge rise in computational cost and memory demand if directly capitalizing on the measure of object relation in \cite{hu2018relation}, not to mention that the increase of supportive object proposals results in more invalid proposals, which may affect the overall stability of relation learning. To alleviate these issues, we propose a new multi-stage module as illustrated in Figure \ref{fig:fig1}(b). Our unique design is to progressively schedule relation distillation. We select object proposals with high objectness scores from all support frames and only augment the features of these proposals with object relation to further distill the relation with respect to proposals in reference frame. Such cascaded means, on one hand, could reduce computation and filter out invalid proposals, and on the other, refine object relation~better.

By consolidating the idea of modeling object relation in spatio-temporal context, we novelly present Relation Distillation Networks (RDN) for boosting video object detection. Specifically, Region Proposal Network (RPN) is exploited to produce object proposals from the reference frame and all the support frames. The object proposals extracted from support frames are packed into supportive pool. The goal of our RDN is to augment the feature of each object proposal in the reference frame by aggregating its relation features over the proposals in the supportive pool. RDN employs multi-stage reasoning structure, which includes basic stage and advanced stage. In the basic stage, RDN capitalizes on all the proposals in the supportive pool to measure relation features measured on both appearance and geometry information. The interactions are explored holistically across all the supportive proposals in this stage irrespective of the validity of proposals. Instead, RDN in the advanced stage nicely selects supportive proposals with high objectness scores and first endows the features of these proposals with relation against all the supportive proposals. Such aggregated features then in turn strengthen the relation distillation with respect to proposals in the reference frame. The upgraded feature of each proposal with object relation is finally exploited for proposal classification and regression. Moreover, the learnt relation also benefit the post-processing of box linking. Note that our RDN is applicable in any region-based vision~tasks.

\begin{figure*}
  \centering
  \vspace{-0.1in}
  \includegraphics[width=0.95\linewidth]{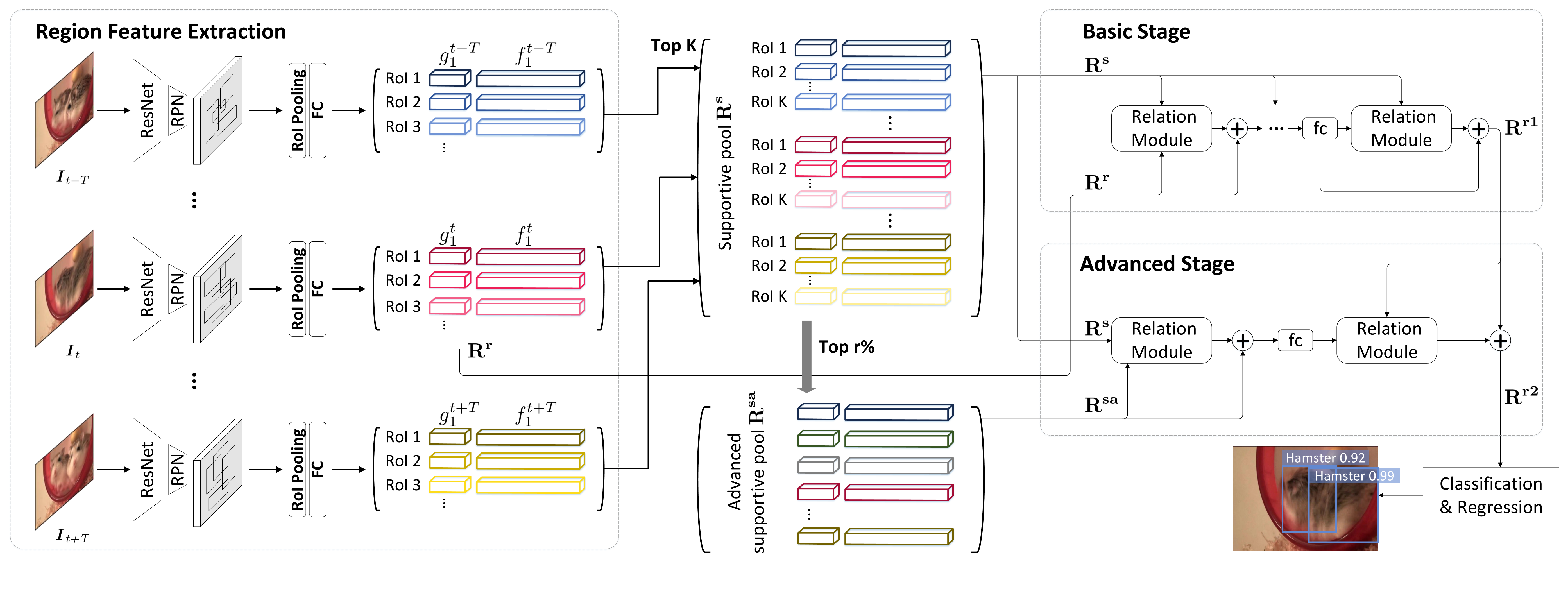}
  \vspace{-0.15in}
  \caption{An overview of Relation Distillation Networks (RDN) for video object detection. Given the input reference frame $I_t$ and all support frames $\{I_\tau\}_{\tau=t-T}^{t+T}$, Region Proposal Networks (RPN) is first employed to produce object proposals (\emph{i.e.}, Region of Interests (RoI)) from reference frame and all support frames. We select the top-$K$ object proposals from reference frame as the reference object set $\mathbf{R^r}$ and pack all the top-$K$ object proposals from support frames into the supportive pool $\mathbf{R^s}$. After that, RDN is devised to augment the feature of each reference proposal in $\mathbf{R^r}$ by aggregating its relation features over the supportive proposals in $\mathbf{R^s}$, enabling the modeling of object relations in spatio-temporal context. Specifically, RDN is a multi-stage module which contains \emph{basic stage} and \emph{advanced stage} to support multi-stage reasoning and relation distillation. In basic stage, all supportive proposals in $\mathbf{R^s}$ are leveraged to measure the relation feature of each reference proposal in $\mathbf{R^r}$ via exploring the interactions across all the supportive proposals, outputting a set of refined reference proposals $\mathbf{R^{r1}}$. In the advanced stage, we first select $r\%$ supportive proposals in $\mathbf{R^s}$ with high objectness scores to form the advanced supportive pool $\mathbf{R^{sa}}$, where the feature of each supportive proposal is endowed with relation against all the supportive proposals. Such aggregated features then in turn strengthen the relation distillation with respect to proposals in $\mathbf{R^{r1}}$ from basic stage. Finally, the upgraded features of all reference proposals ($\mathbf{R^{r2}}$) output from advanced stage is exploited for proposal classification and regression.}
  \label{framework}
  \vspace{-0.15in}
\end{figure*}

\section{Related Work}
\textbf{Object Detection.}
The recent advances in deep convolutional neural networks \cite{he2016resnet,krizhevsky2012imagenet,Simonyan:ICLR15,Szegedy:CVPR15} and well-annotated datasets \cite{lin2014microsoft,ILSVRC15} have inspired the remarkable improvements of image object detection \cite{dai2016r,girshick2015fast,girshick2014rich,he2017mask,he2015spatial,law2018cornernet,lin2017feature,Lin_2017_ICCV,liu2016ssd,redmon2016you,ren2015faster,sermanet2014overfeat}. There are generally two directions for object detection. One is proposal-based two-stage detectors (e.g., R-CNN \cite{girshick2014rich}, Fast R-CNN \cite{girshick2015fast}, and Faster R-CNN \cite{ren2015faster}), and the other is proposal-free one-stage detectors (e.g., SSD \cite{liu2016ssd}, YOLO \cite{redmon2016you}, and RetinaNet \cite{Lin_2017_ICCV}). Recently, motivated by the success of attention model in NLP field \cite{gehring2017convolutional,vaswani2017attention}, \cite{hu2018relation,wang2018non} extend attention mechanisms to support computer vision tasks by exploiting the attention/relations among regions/CNN features. In particular, \cite{hu2018relation} presents an object relation module that models the relations of region proposals through the interaction between their appearance features and coordinates information. \cite{wang2018non} plugs non-local operation into conventional CNN to enable the relational interactions within CNN feature maps, aiming to capture contextual information and eventually boost both object detection and video classification tasks.

The Relation Distillation Networks in our work is also a type of relation modeling among objects. Unlike \cite{hu2018relation} that is developed for object detection in images, ours goes beyond the mining of object relation within one image and aims to explore the object interactions across multiple frames in the complex spatio-temporal context of video object detection. Moreover, a progressive schedule of relation distillation is devised to refine object relations and meanwhile reduce the computational cost on measuring object relations between reference frame and all nearby support frames.

\textbf{Video Object Detection.}
Generalizing still image detectors to video domain is not trivial due to the spatial and temporal complex variations existed in videos, not to mention that the object appearances in some frames may be deteriorated by motion blur or occlusion. One common solution to amend this problem is feature aggregation \cite{stsn,Liu_2018_CVPR,xiao2018matchtrans,zhu2018towards,zhu2017fgfa,zhu2017dff} that enhances per-frame features by aggregating the features of nearby frames. Specifically, FGFA \cite{zhu2017fgfa} utilizes the optical flow from FlowNet \cite{dosovitskiy2015flownet} to guide the pixel-level motion compensation on feature maps of adjacent frames for feature aggregation. \cite{xiao2018matchtrans} devises a spatio-temporal memory module to perform frame-by-frame spatial alignment for aggregation. Another direction of video object detection is box-level association \cite{feichtenhofer2017detect,han2016seq,kang2017object,kang2017t,wang2018manet} which associates bounding boxes from consecutive frames to generate tubelets via independent processes of linking/tracking. Seq-NMS \cite{han2016seq} builds temporal graph according to jaccard overlap between bounding boxes of consecutive frames and searches the optimal path with highest confidence as tubelets. D\&T \cite{feichtenhofer2017detect} integrates a tracking formulation into R-FCN \cite{dai2016r} to simultaneously perform object detection and across-frame track regression. \cite{wang2018manet} further extends FGFA \cite{zhu2017fgfa} by calibrating the object features on box level to boost video object detection.

Despite both feature-level and box-level methods have generally enhanced video object detection with higher quantitative scores, the object relations are not fully exploited across frames for object detection in videos. In contrast, we exploit the modeling of object relations in spatio-temporal context to facilitate video object detection. To this end, we design a novel Relation Distillation Networks to aggregate and propagate object relation across frames to augment object features in a cascaded manner for detection.

\section{RDN for Video Object Detection}
In this paper, we devise Relation Distillation Networks (RDN) to facilitate object detection in videos by capturing the interactions across objects in spatio-temporal context. Specifically, Region Proposal Networks (RPN) is first exploited to obtain the object proposals from the reference frame and all the support frames. RDN then aggregates and propagates object relation over the supportive proposals to augment the feature of each reference object proposal for detection. A multi-stage module is employed in RDN to simultaneously model object relation via multi-stage reasoning and progressively distill relation through refining supportive object proposals with high objectness scores in a cascaded manner. The learnt relation can be further exploited in both classification \& regression for detection and the detection box linking in post-processing. An overview of our RDN architecture is depicted in Figure \ref{framework}.

\subsection{Overview}
\textbf{Notation.}
In the standard task of video object detection, we are given a sequence of adjacent frames $\{{I}_\tau\}_{\tau=t-T}^{t+T}$ where the central frame $I_t$ is set as the reference frame. The whole sequence of adjacent frames $\{{I}_\tau\}_{\tau=t-T}^{t+T}$ is taken as support frames and $T$ represents the temporal spanning range of support frames. As such, the goal of video object detection is to detect objects in reference frame $I_t$ by additionally exploiting the spatio-temporal correlations in the support frames. Since the ultimate goal is to model object relation in spatio-temporal context to boost video object detection, RPN is first leveraged to generate object proposals of reference frame and all support frames. The set of selected top-$K$ object proposals from reference frame is denoted as $\mathbf{R^r} =\{R^r_i\}$. All the top-$K$ object proposals from support frames are grouped as the supportive pool, denoted as $\mathbf{R^s}=\{R^s_i\}$. In addition, we further refine the supportive pool $\mathbf{R^s}$ by sampling $r\%$ supportive object proposals with high objectness scores, leading to the advanced supportive pool $\mathbf{R^{sa}}=\{R^{sa}_i\}$. Both of the supportive pool $\mathbf{R^s}$ and advanced supportive pool $\mathbf{R^{sa}}$ will be utilized in our devised Relation Distillation Networks to enable the progressive scheduling of relation distillation.

\textbf{Problem Formulation.}
Inspired by the recent success of exploring object relations in various vision tasks (e.g., recognition \cite{wang2018videos} and object detection \cite{hu2018relation}), we formulate our video object detection method by modeling interactions between objects in spatio-temporal context to boost video object detectors. Given the set of reference proposals $\mathbf{R^r}$, the supportive pool $\mathbf{R^s}$ and the advanced supportive pool $\mathbf{R^{sa}}$, we are interested to progressively augment the feature of each reference proposal in $\mathbf{R^r}$ with distilled relations against supportive proposals in $\mathbf{R^s}$ and $\mathbf{R^{sa}}$. To do this, a novel Relation Distillation Networks is built based on the seminal detector Faster R-CNN \cite{ren2015faster}. A multi-stage reasoning structure consisting of basic and advanced stages is adopted in RDN for progressively scheduling relation distillation in a cascaded manner. Such design of cascaded means not only reduces computation and filters out invalid proposals, but also progressively refines object relations of reference proposals against supportive ones to boost detection. Most specifically, in the basic stage, all supportive proposals in $\mathbf{R^s}$ are utilized to measure relation features of reference proposals in $\mathbf{R^r}$ on both appearance and geometry information. As such, the output set of refined reference proposals $\mathbf{R^{r1}}=\{R^{r1}_i\}$ from basic stage is obtained via a stacked relation module which explores the interactions between reference proposals and all supportive proposals irrespective of the validity of proposals. In the advanced stage, we first enhance the feature of each selected supportive proposal in the advanced supportive pool $\mathbf{R^{sa}}$ with relation against all the supportive proposals in $\mathbf{R^s}$. Such aggregated features of distilled supportive proposals then in turn strengthen the relation distillation with respect to reference proposals in $\mathbf{R^{r1}}$ output from basic stage. Once the upgraded reference proposals $\mathbf{R^{r2}}=\{R^{r2}_i\}$ from advance stage are obtained, we directly exploit them to improve object detection in reference frame. More details about the multi-stage reasoning structure of our RDN will be elaborated in Section \ref{sec.architecture}. Moreover, by characterizing the natural interactions between objects across frames, the learnt relations can be further leveraged to guide detection box linking in post-processing, which will be presented in Section \ref{sec.linking}.

\subsection{Object Relation Module}\label{sec.background}
We begin by briefly reviewing object relation module \cite{hu2018relation} for object detection in images. Motivated from Multi-Head Attention in \cite{vaswani2017attention}, given the input of proposals $\mathbf{R}=\{R_i\}$, object relation module is devised to enhance each proposal $R_i$ by measuring $M$ relation features as the weighted sum of appearance features from other proposals. Note that we represent each object proposal with its geometric feature $g_i$ (\emph{i.e.}, the 4-dimensional coordinates of object proposal) and appearance feature $f_i$ (\emph{i.e.}, the RoI pooled feature of object proposal). Formally, the $m$-th relation feature of proposal $R_i$ is calculated conditioning on $\mathbf{R}$:
\begin{equation}\label{relation_feature}\small
  f_{rela}^{m}(R_i,\mathbf{R}) = \sum\nolimits_{j}\omega_{ij}^m\cdot(W_L^m\cdot{f}_j),~~~m=1,\cdot\cdot\cdot,M,
\end{equation}
where $W^m_L$ denotes the transformation matrix. $\omega_{ij}$ is an element in relation weight matrix ${\bf{\omega}}$ and represents the pairwise relation between proposals $R_i$ and $R_j$ which is measured based on their appearance and geometric features. By concatenating all the $M$ relation features of each proposal $R_i$ and its appearance feature, we finally obtain the relation-augmented feature output from object relation module:
\begin{equation}\label{relation_out}\small
  f_{rm}(R_i,\mathbf{R}) = f_i + concat[\{f_{rela}^{m}(R_i,\mathbf{R})\}_{m=1}^{M}].
\end{equation}

\subsection{Relation Distillation Networks}\label{sec.architecture}
Unlike \cite{hu2018relation} that explores object relations within an image for object detection, we facilitate the modeling of object relations in video object detection by exploiting the object interactions across multiple frames under the complex spatio-temporal context. One natural way to extend the relation-augmented detector in image to video is to capitalize on the object relation module in \cite{hu2018relation} to measure the interactions between the objects in reference frame and all supportive objects from nearby frames. Nevertheless, such way will lead to a huge rise in computational cost, not to mention that the increase of supportive proposals results in more invalid proposals and the overall stability of relation learning will be inevitably affected. To alleviate this issue, we devise Relation Distillation Networks to progressively schedule relation distillation for enhancing detection via a multi-stage reasoning structure, which contains basic stage and advanced stage. The spirit behind follows the philosophy that basic stage explores relations holistically across all the supportive proposals with respect to reference proposals, and advanced stage progressively distills relations via refining supportive proposals, which are augmented with relations to further strengthen reference proposals.

\textbf{Basic Stage.} Formally, given the set of reference proposals $\mathbf{R^r}$ and the supportive pool $\mathbf{R^s}$, the basic stage predicts the relation features of each reference proposal as the weighted sum of features from all supportive proposals via a stacked relation module:
\begin{equation}\label{equ:basic}\small
  \mathbf{R^{r1}} = \mathcal{N}_{basic}(\mathbf{R^r},\mathbf{R^s}),
\end{equation}
where $\mathcal{N}_{basic}\left( \cdot \right)$ represents the function of the stacked relation module in basic stage and $\mathbf{R^{r1}}$ denotes the output enhanced reference proposals from basic stage. Please note that in the complex spatio-temporal context of video object detection, a single relation module is insufficient to model the interactions between objects among multiple frames. Therefore, we iterate the relation reasoning in a stacked manner equipped with $N_b$ object relation modules to better characterize the relations across all the supportive proposals with regard to reference proposals. Specifically, for the $k$-th object relation module in basic stage, the $i$-th reference proposal is augmented with the relation features over all proposals in supportive pool $\mathbf{R^s}$:
\begin{equation}\label{equ:basic_detail}\small
  R^{r1,k}_i  = \begin{cases}
  f_{rm}(R^r_i,\mathbf{R^s}), & k=1,\\
  f_{rm}(h({R}^{r1,k-1}_i),\mathbf{R^s}), & k>1,
  \end{cases}
\end{equation}
where $h\left( \cdot \right)$ denotes the feature transformation function implemented with a fully-connected layer plus ReLU. Each relation module takes the transformed features of reference proposals from previous relation module as the reference inputs. We stack $N_b$ relation modules in basic stage and all the enhanced reference proposals from the $N_b$-th relation module are taken as the output $\mathbf{R^{r1}}$ of basic stage.

\textbf{Advanced Stage.}
The relation reasoning in basic stage only explores the modeling of interactions between reference proposal and all the supportive proposals, while leaving the relations among supportive proposals in $\mathbf{R^s}$ unexploited. Furthermore, we present a novel advanced stage to explore the interactions between supportive proposals by enhancing the distilled supportive proposals with relations against all supportive proposals. Next, the enhanced distilled supportive proposals are utilized to further strengthen the reference proposals from basic stage via relation reasoning in between. Such design of progressively distilling supportive proposals in advanced stage not only reduces the computation cost of measuring relations among supportive proposals, but also filters out invalid supportive proposals for relation reasoning and eventually improves detection.

\begin{algorithm}[t]
  \caption{Inference Algorithm of our RDN}
  \small
  \begin{algorithmic}[1] % [1] for line numbers
  \State \textbf{Input}: video frames $\{I_t\}$, temporal spanning range $T$.
  \For{$t=1$ \textbf{to} $T+1$}                            \Comment{initialize proposal feature buffer}
  \State $\mathbf{R}_t = \mathcal{N}_{RoI}(I_t)$                   \Comment{region proposal and feature extraction}
  \State $\mathbf{R}^{\mathbf{s}}_t = \mathbf{Sample}_{top{\text -}K}(\mathbf{R}_t)$  \Comment{sample top-$K$ proposals}
  \EndFor

  \For{$t=1$ \textbf{to} $\infty$}		
  \State $\mathbf{R^r} = \mathbf{R}_t$                      \Comment{reference proposal set}
  \State $\mathbf{R^s} = \mathbf{R}^\mathbf{s}_{max(1,t\text{-}T)}\cup {\cdot\cdot\cdot} \cup\mathbf{R}^\mathbf{s}_{t\text{+}T}$  \Comment{supportive pool}
  \State $\mathbf{R^{r1}} = \mathcal{N}_{basic}(\mathbf{R^r},\mathbf{R^s})$ \Comment{basic stage}
  \State $\mathbf{R^{sa}} = \mathbf{Sample}_{top{\text -}r\%}(\mathbf{R^s})$     \Comment{sample top-$r\%$ proposals}
  \State $\mathbf{R^{r2}} = \mathcal{N}_{adv}(\mathbf{R^{r1}}, \mathbf{R^s}, \mathbf{R^{sa}})$ \Comment{advanced stage}
  \State $\mathbf{D}_t = \mathcal{N}_{det}(\mathbf{R^{r2}})$  \Comment{classification and regression}
  \State $\mathbf{R}_{t\text{+}T\text{+}1} = \mathcal{N}_{RoI}(I_{t\text{+}T\text{+}1})$
  \State $\mathbf{R}^{\mathbf{s}}_{t\text{+}T\text{+}1} = \mathbf{Sample}_{top{\text -}K}(\mathbf{R}_{t\text{+}T\text{+}1})$
  \State \textbf{update proposal feature buffer}

  \EndFor
  \State \textbf{Output}: detection results $\{\mathbf{D}_{t}\}$
  \end{algorithmic}
  \label{alg.RDN_inference}
  \end{algorithm}

Technically, given the output reference proposals $\mathbf{R^{r1}}$ from basic stage, the supportive pool $\mathbf{R^s}$, and the advanced supportive pool $\mathbf{R^{sa}}$, the advanced stage further strengthens all reference proposals $\mathbf{R^{r1}}$ through progressively scheduling relation distillation:
\begin{equation}\label{equ:advance}\small
  \mathbf{R^{r2}} = \mathcal{N}_{adv}(\mathbf{R^{r1}},\mathbf{R^s},\mathbf{R^{sa}}),
\end{equation}
where $\mathcal{N}_{adv}\left( \cdot \right)$ denotes the operation in advanced stage and $\mathbf{R^{r2}}$ represents the output relation-augmented reference proposals from advanced stage.
Most specifically, we first refine the distilled supportive proposals in $\mathbf{R^{sa}}$ with relation reasoning against all supportive proposals in $\mathbf{R^s}$:
\begin{equation}\label{equ:advance_1}\small
  R^{a}_i = f_{rm}(R^{sa}_i,\mathbf{R^s}),
\end{equation}
where $R^{a}_i$ denotes the $i$-th refined supportive proposal. After that, the refined supportive proposals $\mathbf{R^a} =\{R^{a}_i\}$ are utilized to further distill the relation with respect to reference proposals $\mathbf{R^{r1}}$ from basic stage:
\begin{equation}\label{equ:advance_2}\small
R^{r2}_i = f_{rm}(R^{r1}_i,\mathbf{R^a}),
\end{equation}
where $R^{r2}_i$ denotes the $i$-th upgraded reference proposal. Finally, all the upgraded reference proposals $\mathbf{R^{r2}}=\{R^{r2}_i\}$ are exploited for proposal classification and regression.

\textbf{Training and Inference.}
At training stage, we adopt the strategy of temporal dropout \cite{zhu2017fgfa} to randomly select two support frames $I_{t+\tau_1}$ and $I_{t+\tau_2}$ ($\tau_1,\tau_2\in[-T, T]$) from the adjacent frames $\{{I}_\tau\}_{\tau=t-T}^{t+T}$. Accordingly, the whole RDN is optimized with both classification and regression losses over the relation-augmented reference proposals $\mathbf{R^{r2}}$ from advanced stage in an end-to-end manner.

During inference, we follow \cite{zhu2017fgfa} and sequentially process each frame with a sliding proposal feature buffer of adjacent frames $\{{I}_\tau\}_{\tau=t-T}^{t+T}$. The capacity of this proposal feature buffer is set as the length of adjacent frames (\emph{i.e}., $2T+1$), except for the beginning and ending $T$ frames. The detailed inference process of RDN is given in Algorithm \ref{alg.RDN_inference}.

\subsection{Box Linking with Relations}\label{sec.linking}

To further boost video object detection results by re-scoring individual detection boxes among consecutive frames, we adopt the post-processing of linking detection boxes across frames as in \cite{gkioxari2015finding,han2016seq,kang2017t}. Despite the box-level post-processing methods have generally enhanced video object detection with higher quantitative scores, the object relations between detection boxes are not fully studied for box linking. In contrast, we integrate the learnt object-to-object relations into post-processing of box linking to further propagate the confidence scores among high-related detection boxes and thus improve the detection.

Specifically, we formulate the post-processing of box linking as an optimal path finding problem. Note that since the box linking is independently applied for each class, we omit the notation of class here for simplicity. Given two detection boxes $d_i^t$ and $d_j^{t+1}$ from consecutive frames $I_t$ and $I_{t+1}$, the linking score between them is calculated as:
\begin{equation}\label{link_score}\small
  S(d_i^t, d_j^{t+1}) = \{s_i^t + s_j^{t+1} + iou(d_i^t, d_j^{t+1})\} \cdot e^{\bar{\omega}_{ij}},
\end{equation}
where $s_i^t$ and $s_j^{t+1}$ are confidence scores of the two boxes, and $iou(\cdot)$ indicates jaccard overlap. $\bar{\omega}_{ij}$ represents the pairwise relation weight between the two boxes $d_i^t$ and $d_j^{t+1}$, which is measured as the average of all the $M$ relation weights obtained in the last relation module at basic stage: $\bar{\omega}_{ij} = \frac{1}{M}\sum_{m=1}^M \omega_{ij} ^m$. Accordingly, for each class, we seek the optimal path as:
\begin{equation}\small
  \bar{P}^\ast=\mathop{\arg\max}_{\bar{P}} \frac{1}{\mathcal{L}} \sum_{t=1}^{\mathcal{L}-1} S(\mathbf{D}_t, \mathbf{D}_{t+1})
  \label{eq:viterbi},
  \end{equation}
where $\mathbf{D}_t = \{d^t_i\}$ denotes the set of detection boxes in frame $I_t$ and $\mathcal{L}$ is the duration of video. This problem can be solved by Viterbi algorithm \cite{gkioxari2015finding}. Once the optimal path for linking boxes is obtained, we follow \cite{feichtenhofer2017detect} and re-score detection boxes in each tube by adding the average value of the top-50\% classification score of boxes in this path.
%------------------------------------------------------------------------
\section{Network Architecture}
\textbf{Backbone.} We exploit two kinds of backbones, \emph{i.e.}, ResNet-101 \cite{he2016resnet} and ResNeXt-101-64$\times$4d \cite{xie2017aggregated}, for our RDN. Specifically, to enlarge the resolution of feature maps, we modify the stride of first conv block in last stage of convolutional layers from 2 to 1. As such, the effective stride in this stage is changed from 32 to 16 pixels. Besides, all the 3$\times$3 conv layers in this stage are modified by the ``hole algorithm" \cite{chen2014semantic,mallat1999wavelet} (\emph{i.e.}, ``atrous convolution" \cite{long2015fully}) to compensate for the receptive fields.

\textbf{Region Feature Extraction.} We utilize RPN \cite{ren2015faster} on the top of conv4 stage for region feature extraction. In particular, we leverage 12 anchors with 4 scales $\{64^2, 128^2, 256^2, 512^2\}$ and 3 aspect ratios $\{$1$:$2,1$:$1,2$:$1$\}$ for classification and regression. During training and inference, we first pick up 6,000 proposals with highest objectness scores and then adopt Non Maximum Suppression (NMS) with threshold of 0.7 Intersection-over-Union (IoU) to obtain $N=300$ proposals for each frame. After generating region proposals, we apply RoI pooling followed by a 1,024-d fully-connected layer on the top of conv5 stage to extract RoI feature of each proposal.

\textbf{Relation Distillation Networks.} For each relation module in RDN, the number of relation features is set as $M=16$. The dimension of each relation feature is 64. As such, by concatenating all the $M=16$ relation features as in Equation \ref{relation_out}, the dimension of relation-augmented feature output from relation module is 1,024. In basic stage, we stack $N_b=2$ relation modules. In advanced stage, one relation module is first employed to enhance proposals in advanced supportive pool $\mathbf{R^{sa}}$. Next we apply another relation module to strengthen the reference proposals output from basic stage. Finally, we utilize two parallel branches (\emph{i.e.}, classification and regression) to obtain detection boxes based on the refined RoI features from advanced stage.

\section{Experiments}

\subsection{Dataset and Evaluation}
We empirically verify the merit of our RDN by conducting experiments on ImageNet object detection from video (VID) dataset \cite{ILSVRC15}. The ImageNet VID dataset is a large-scale benchmark for video object detection task, consisting of 3,862 training videos and 555 validation videos from 30 classes. Given the fact that the ground truth of the official testing set are not publicly available, we follow the widely adopted setting as in \cite{feichtenhofer2017detect,kang2017object,wang2018manet,xiao2018matchtrans,zhu2017fgfa,zhu2018towards} to report mean Average Precision (mAP) on validation set.

Following the common protocols in \cite{feichtenhofer2017detect,wang2018manet,xiao2018matchtrans,zhu2017fgfa}, we utilize both ImageNet VID and ImageNet object detection (DET) dataset to train our RDN. Since the 30 classes in ImageNet VID are a subset of 200 classes in ImageNet DET dataset, we adopt the images from overlapped 30 classes in ImageNet DET for training. Specifically, due to the redundancy among adjacent frames, we sample 15 frames from each video in ImageNet VID for training. For ImageNet DET, we select at most 2,000 images from each class to make the class distribution more balanced.

\begin{table}
  \centering
  %\small
  \footnotesize
  \caption{Performance comparison with state-of-the-art end-to-end video object detection models on ImageNet VID validation set.}
  \setlength{\tabcolsep}{1.1mm}
  \begin{tabular}{l|c|c|c}
  \Xhline{2\arrayrulewidth}
  Methods                           & Backbone                                        &Base Detector        & mAP (\%)      \\
  \hline
  \hline
  \multirow{2}{*}{FGFA\,\cite{zhu2017fgfa}}  &ResNet-101             &R-FCN           & 76.3          \\
                                             &           ResNet-101                             &Faster R-CNN    & 77.5          \\
  \hline
  MANet\,\cite{wang2018manet}       & ResNet-101                                      &R-FCN           & 78.1          \\
  \hline
  THP\,\cite{zhu2018towards}        & ResNet-101\,+\,DCN\,\cite{dai2017deformable}    &R-FCN           & 78.6          \\
  \hline
  STSN\,\cite{stsn}                 & ResNet-101\,+\,DCN\,\cite{dai2017deformable}    &R-FCN           & 78.9          \\

  \Xhline{2\arrayrulewidth}
  \multirow{2}{*}{RDN}            & ResNet-101                                        &Faster R-CNN    &$\mathbf{81.8}$ \\
                                   & ResNeXt-101-64$\times$4d                                &Faster R-CNN    &$\mathbf{83.2}$ \\
  \Xhline{2\arrayrulewidth}
  \end{tabular}
  \vspace{-0.25in}
  \label{tab.comp.feature_enhance}
  \end{table}

\subsection{Implementation Details}
At training and inference stages, the temporal spanning range is set as $T=18$. We select the top $K=75$ proposals with highest objectness scores from each support frame and pack them into the supportive pool $\mathbf{R^s}$. We obtain the advanced supportive pool $\mathbf{R^{sa}}$ by sampling 20\% supportive proposals with highest objectness scores from $\mathbf{R^s}$.

We implement RDN mainly on Pytorch 1.0 \cite{paszke2017automatic}. The input images are first resized so that the shorter side is 600 pixels. The whole architecture is trained on four Tesla V100 GPUs with synchronized SGD (momentum: 0.9, weight decay: 0.0001). There is one mini-batch in each GPU and each mini-batch contains one image/frame. For reference frame, we sample 128 RoIs with a ratio of 1:3 for positive:negatives. We adopt a two-phase strategy for training our RDN. In the first phase, we train the basic stage together with backbone \& RPN over the combined training set of ImageNet VID and ImageNet DET for 120k iterations. The learning rate is set as 0.001 in the first 80k iterations and 0.0001 in the next 40k iterations. In the second phase, the whole RDN architecture is trained on the combined training set with another 60k iterations. The learning rate is set as 0.001 in the first 40k iterations and 0.0001 in the last 20k iterations. The whole procedure of training takes about 15 hours in the first phase and 8 hours in the second phase. At inference, we adopt NMS with a threshold of 0.5 IoU to suppress reduplicate detection boxes.

\subsection{Performance Comparison}
\textbf{End-to-End models.}
The performances of different end-to-end video object detection models on ImageNet VID validation set are shown in Table \ref{tab.comp.feature_enhance}. Note that for fair comparison, here we only include the state-of-the-art end-to-end techniques which purely learn video object detector by enhancing per-frame feature in an end-to-end fashion without any post-processing. Overall, the results under the same backbone demonstrate that our proposed RDN achieves better performance against state-of-the-art end-to-end models. In particular, the mAP of RDN can achieve 81.8\% with ResNet-101, which makes 2.9\% absolute improvement over the best competitor STSN. As expected, when equipped with a stronger backbone (ResNeXt-101-64$\times$4d), the mAP of our RDN is further boosted up to 83.2\%. By additionally capturing global motion clues to exploit instance-level calibration, MANet exhibits better performance than FGFA that performs pixel-level calibration with the guidance from optical flow. Different from the flow-guided methods (FGFA, MANet, and THP) which estimate the motion across frames for warping the feature map, STSN spatially samples features from adjacent frames for feature aggregation and achieves better performance. Nevertheless, the performance of STSN is still lower than that of our RDN which models object relation in spatio-temporal context. The results highlight the advantage of aggregating and propagating object relation to augment object features for video object detection.

\begin{table}
  \centering
  %\small
  \footnotesize
  \caption{Performance comparison with state-of-the-art video object detection methods plus post-processing on ImageNet VID validation set. BLR: Box Linking with Relations in Section \ref{sec.linking}.}
  \setlength{\tabcolsep}{0.666mm}
  \begin{tabular}{l|c|c|c}
  \Xhline{2\arrayrulewidth}
  Methods                                     & Backbone     & Base Detector          & mAP (\%)      \\
  \hline
  \hline
  T-CNN\,\cite{kang2017t}                       & DeepID\,\cite{ouyang2014deepid}\,+\,Craft\,\cite{yang2016craft}  &R-CNN          & 73.8          \\
  \hline
  \multirow{2}{*}{FGFA\,\cite{zhu2017fgfa}\,+\,\cite{han2016seq}}          & ResNet-101 &R-FCN & 78.4         \\

          & Aligned Inception-ResNet & R-FCN & 80.1         \\
  \hline
  \multirow{3}{*}{D\&T\,\cite{feichtenhofer2017detect}}     & ResNet-101  &R-FCN        & 79.8 \\
                                    &    ResNet-101                &Faster R-CNN & 80.2 \\
                                    & Inception-v4    &R-FCN  & 82.0 \\
  \hline
  STMN\,\cite{xiao2018matchtrans}   & ResNet-101     &R-FCN                               & 80.5          \\

  \Xhline{2\arrayrulewidth}

  \multirow{2}{*}{RDN\,+\,\cite{han2016seq} }         & ResNet-101    &Faster R-CNN        & 82.6 \\
                                            & ResNeXt-101-64$\times$4d &Faster R-CNN    & 83.9 \\
  \hline
  \multirow{2}{*}{RDN\,+\,\cite{gkioxari2015finding} }           & ResNet-101    &Faster R-CNN        & 83.4 \\
                                            & ResNeXt-101-64$\times$4d &Faster R-CNN    & 84.5 \\
  \hline
  \multirow{2}{*}{RDN\,+\,BLR}               & ResNet-101       &Faster R-CNN     & $\mathbf{83.8}$ \\
                                            & ResNeXt-101-64$\times$4d &Faster R-CNN    & $\mathbf{84.7}$ \\
  \Xhline{2\arrayrulewidth}
  \end{tabular}
  \vspace{-0.25in}
  \label{tab.comp.box_association}
  \end{table}

\textbf{Add Post-Processing.}
In this section, we compare our RDN with other state-of-the-art methods by further applying post-processing of box linking. Table \ref{tab.comp.box_association} summarizes the results on ImageNet VID validation set. In general, when equipped with existing post-processing techniques (Seq-NMS and Tube Linking), our RDN exhibits better performances than other state-of-the-art post-processing based approaches. In addition, by leveraging our Box Linking with Relations (BLR) that integrates the learnt object relations into Tube Linking, the performances of RDNs are further boosted up to 83.8\% and 84.7\% with ResNet-101 and ResNeXt-101-64$\times$4d, respectively. This confirms the effectiveness of propagating confidence scores among detection boxes with high relations via box linking in our BLR.

\subsection{Experimental Analysis}

\begin{table}[!tb] \footnotesize
  \centering
  \caption{\small Performance comparisons across different ways on the measure of object relation, \emph{i.e.}, Faster R-CNN on single frame irrespective of relation, stacked relation within frame in \cite{hu2018relation}, RDN with relation only in basic stage (\textbf{BASIC}), full version of RDN with advanced stage (\textbf{ADV})). The backbone is ResNet-101.}
  \setlength{\tabcolsep}{4mm}
  \begin{tabular}{l|cc|l}
  \Xhline{2\arrayrulewidth}
  Methods  & BASIC         & ADV          & mAP (\%)                       \\
  \hline
  \hline
  Faster R-CNN                        &               &              & $75.4$                         \\
  \ ~~~~~+~Relation \cite{hu2018relation}  &               &              & $78.5_{\uparrow 3.1}$           \\
  \hline
  Faster R-CNN                        &               &              &              \\
  \ ~~~~~+~BASIC                           & $\checkmark$  &              & $80.9_{\uparrow 5.5}$    \\
  \Xhline{2\arrayrulewidth}
  RDN                                 & $\checkmark$  & $\checkmark$ & $\mathbf{81.8_{\uparrow 6.4}}$ \\
  \Xhline{2\arrayrulewidth}
  \end{tabular}
  \label{tab.ablation.arch}
  \vspace{-0.10in}
\end{table}

\begin{figure}[!tb]
%\vspace{-0.05in}
\centering {\includegraphics[width=0.49\textwidth]{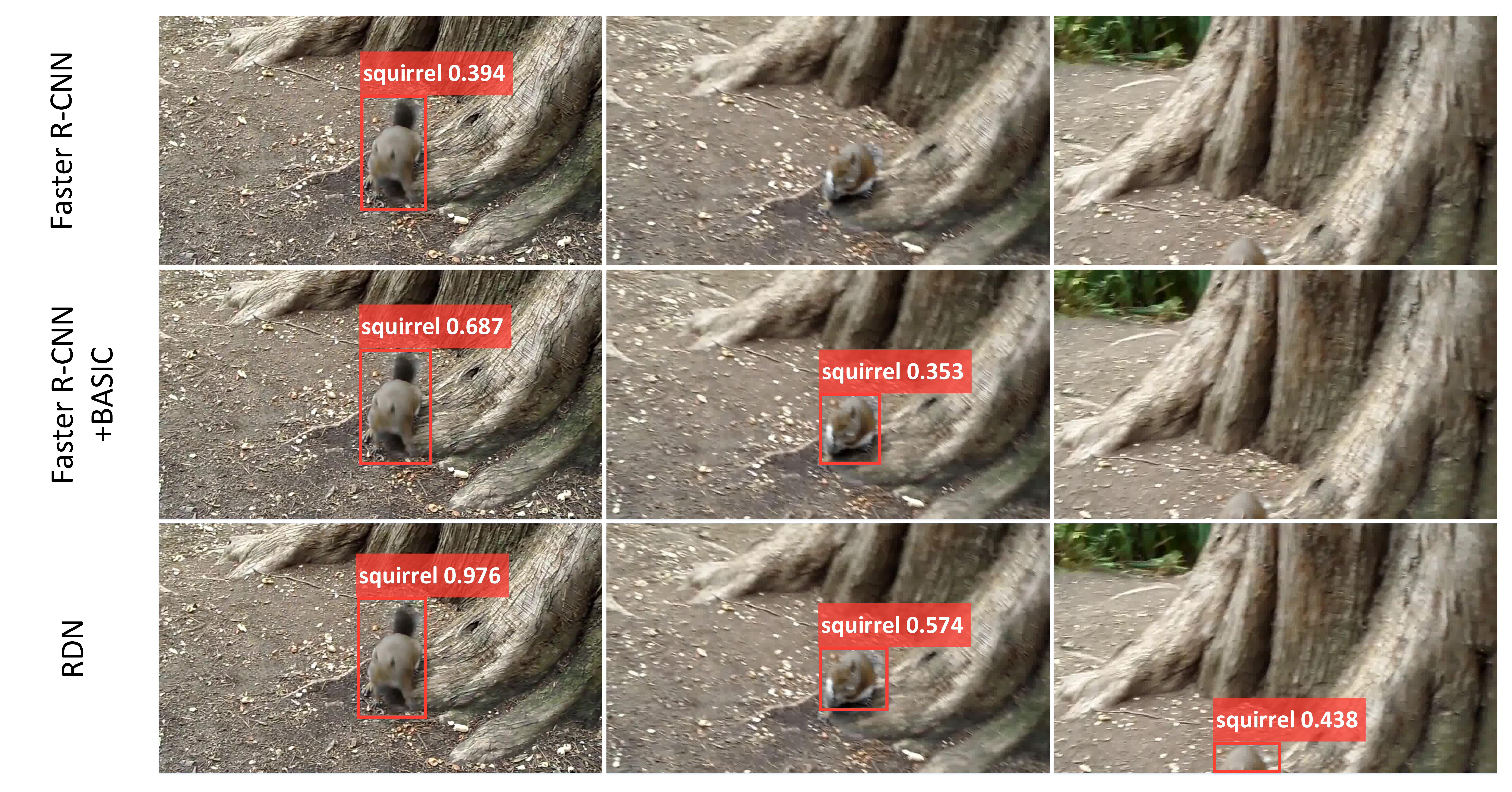}}
\vspace{-0.1in}
\caption{Examples of video object detection results by different ways of relation modeling in our RDN.}
\label{fig.example}
\vspace{-0.15in}
\end{figure}

\textbf{Ablation Study.} Here we study how each design in our RDN influences the overall performance. Faster R-CNN \cite{ren2015faster} simply executes object detection on single frame irrespective of object relation. \cite{hu2018relation} models relation in an image via stacked relation modules. We extend this idea to learn the interactions between objects in a video frame and re-implement \cite{hu2018relation} in our experiments. The run of Faster R-CNN + BASIC only exploits the basic stage for relation reasoning and RDN further integrates the advanced stage.

Table \ref{tab.ablation.arch} details the performances across different ways on the measure of object relation. Directly performing Faster R-CNN on single frame achieves 75.4\% of mAP. The mining of relation in \cite{hu2018relation} leads to a boost of 3.1\%. The results verify the idea of exploring object relation to improve video object detection, even in case when the relation is measured within each frame. By capturing object interactions across frames in the basic stage, Faster R-CNN + BASIC boosts up the mAP from 75.4\% to 80.9\%. The improvements indicate learning relation in spatio-temporal context is superior to spatial dimension only. RDN is benefited from the mechanism of cascaded relation distillation in advanced stage and the mAP of RDN finally reaches 81.8\%. Figure \ref{fig.example} showcases one example of video object detection results with different ways of relation modeling in our RDN. As illustrated in the figure, the detection results become increasingly robust as more designs of relation modeling are included.

\textbf{Effect of Temporal Spanning Range $T$.}
To explore the effect of temporal spanning range $T$ in our RDN, we show the performance and run time by varying this number from 3 to 24 within an interval of 3 in Table \ref{tab.ablation.range}. The best performance is attained when the temporal spanning range is set to $T=18$. In particular, once the temporal spanning range is larger than 12, the performances are less affected with the change of the temporal spanning range, which eases the selection of the temporal spanning range in our RDN practically. Meanwhile, enlarging the temporal spanning range generally increases run time at inference. Thus, the temporal spanning range is empirically set to 18, which is a good tradeoff between performance and run time.

\textbf{Effect of Relation Module Number $N_b$ in Basic Stage.} Table \ref{tab.ablation.stack_Nb} shows the performances of employing different number of relation module in basic stage. In the extreme case of $N_b=0$, no relation module is utilized and the model degenerates to Faster R-CNN on single frame. With the use of only one relation module, the mAP is increased from 75.4\% to 79.4\%. That basically validates the effectiveness of modeling relation for object detection. The mAP is further boosted up to 80.9\% with the design of two modules but the performance slightly decreases when stacking more modules. We speculate that this may be the result of unnecessary information repeat from support frames and that double proves the motivation of designing the advanced stage in RDN. In practice, the number $N_b$ is generally set to 2.

\begin{table}
  \centering
  \footnotesize
  \vspace{-0.00in}
  \caption{Performance and run time comparisons by using different temporal spanning range $T$ in our RDN.}
  \setlength{\tabcolsep}{1.2mm}
  \begin{tabular}{l|c|c|c|c|c|c|c|c}
  \Xhline{2\arrayrulewidth}
  \# $T$ & 3    & 6    & 9    & 12  & 15   & 18  & 21   &  24    \\
  \hline
  \hline
  mAP (\%)     & 80.3 & 80.7 & 80.9 & 81.3 & 81.6 & $\mathbf{81.8}$ & 81.7 & 81.7 \\
  \hline
  runtime (ms) & 90.1 & 90.3 & 91.5 &  93.0    &  93.5    & 94.2 & 97.3   &103.1\\
  \Xhline{2\arrayrulewidth}
  \end{tabular}
  \vspace{-0.05in}
  \label{tab.ablation.range}
  \end{table}

\begin{table}
  \centering
  \footnotesize
  \vspace{-0.00in}
  \caption{Performance comparisons by using different number of relation module in basic stage.}
  \setlength{\tabcolsep}{3.2mm}
  \begin{tabular}{l|c|c|c|c|c}
  % \hline
  %  \# training frames & \multicolumn{7}{c|}{ 2*} & \multicolumn{7}{c}{ 5} \\
  \Xhline{2\arrayrulewidth}
  \# $N_b$ & 0 & 1 & 2 & 3 & 4 \\
  \hline
  \hline
  mAP (\%) & 75.4 & 79.4 & $\mathbf{80.9}$ & 80.8 & 80.4\\
  \Xhline{2\arrayrulewidth}
  \end{tabular}
  \vspace{-0.15in}
  \label{tab.ablation.stack_Nb}
  \end{table}

\textbf{Effect of Sampling Number $K$ in Basic Stage and Sampling Ratio $r$\% in Advanced Stage.}
We firstly vary $K$ from 25 to 300 in basic stage to explore the relationship between the performance/run time and the sampling number $K$. As shown in Table \ref{tab.ablation.k}, the performances are very slightly affected with the change of sampling number $K$. Specifically, the best performance is attained when the sampling number $K$ is 75. Meanwhile, the run time at inference is gradually increased when enlarging the sampling number. Therefore, we set the sampling number $K$ to 75 practically. Next, to investigate the effect of sampling ratio $r$\% in advanced stage, we further compare the results of performance and run time by varying the sampling ratio from 10\% to 100\% in Table \ref{tab.ablation.ratio}. The best performance is obtained when the sampling ratio is set as $20$\%. Meanwhile, the performances are relatively smooth when the sampling ratio varies. That practically eases the selection of of the sampling ratio $r$\% in advanced stage. In addition, when the sampling ratio increases, the run time is significantly increased. Accordingly, the sampling ratio is empirically set as $r=20$\%, which seeks a better tradeoff between performance and run time.

%\textbf{Effect of Sampling Ratio $r$\% in Advanced Stage.}
%In advanced stage, we obtain the advanced supportive pool by sampling $r=20$\% proposals with highest objectness scores from supportive pool. Next, to investigate the effect of sampling ratio $r$\%, we compare the results of performance and run time by varying the sampling ratio from 10\% to 100\% in Table \ref{tab.ablation.ratio}. The best performance is obtained when the sampling ratio is set as $20$\%. Meanwhile, the performances are relatively smooth when the sampling ratio varies. That practically eases the selection of of the sampling ratio $r$\% in advanced stage. In addition, when the sampling ratio increases, the run time is significantly increased. Accordingly, the sampling ratio is empirically set as $r=20$\%, which seeks a better tradeoff between performance and run time.

\begin{table}
  \centering
  \footnotesize
  \vspace{-0.00in}
  \caption{Performance and run time comparisons by using different sampling number $K$ in basic stage of our RDN.}
  \setlength{\tabcolsep}{1.2mm}
  \begin{tabular}{l|c|c|c|c|c|c|c|c}
  \Xhline{2\arrayrulewidth}
  \# $K$ & 25    & 50    & 75    & 100    & 150  & 200   &  250&  300    \\
  \hline
  \hline
  mAP (\%)     & 80.2& 80.5& 80.9& 80.7&80.4& 80.4& 80.2& 80.1 \\
  \hline
  runtime (ms) & 80.0& 81.7& 84.9& 86.3&  94.9& 107.2& 125.3& 152.7\\
  \Xhline{2\arrayrulewidth}
  \end{tabular}
  \vspace{-0.05in}
  \label{tab.ablation.k}
  \end{table}

\begin{table}
  \centering
  \footnotesize
  \vspace{-0.00in}
  \caption{Performance and run time comparisons by using different sampling ratio $r$\% in advanced stage of our RDN.}
  \setlength{\tabcolsep}{1.2mm}
  \begin{tabular}{l|c|c|c|c|c|c|c|c}
  \Xhline{2\arrayrulewidth}
  \# $r$ (\%) &10& 20& 30& 40& 50& 60& 80& 100   \\
  \hline
  \hline

  mAP (\%)    &   81.3& 81.8& 81.7& 81.6& 81.5& 81.5& 81.3& 81.3\\
  \hline
  runtime (ms)  & 92.8& 94.2& 96.9& 100.2& 104.0& 108.9& 114.8& 125.6\\
  \Xhline{2\arrayrulewidth}
  \end{tabular}
  \vspace{-0.25in}
  \label{tab.ablation.ratio}
  \end{table}

\textbf{Complementarity of Two Stages.}
In RDN, basic stage augments reference proposals with relation features of supportive proposals, which enhances reference proposals with first-order relation from supportive proposals on a star-graph. Then advanced stage progressively samples supportive proposals with high objectness scores and first enhances sampled/advanced proposals with relation against all supportive proposals. In this way, the advanced supportive proposals is endowed with first-order relation from supportive proposals on a full-connected graph. Next, advanced stage strengthens reference proposals with advanced supportive proposals. As such, the reference proposals are further endowed with higher-order relation from supportive proposals, which are naturally complementary
to basic stage.

\section{Conclusions}

We have presented Relation Distillation Networks architecture, which models object relation across frames to boost video object detection. Particularly, we study the problem from the viewpoint of employing multi-stage reasoning and scheduling relation distillation progressively. To verify this, we utilize RPN to generate object proposals in the reference and support frames. The supportive pool is comprised of all the proposals extracted from support frames. In the basic stage, RDN measures the relation of each object proposal in reference frame over all the proposals in the supportive pool and augments the features with relation. In the advanced stage, RDN self adjusts the selected supportive proposals with the relation against all the supportive ones firstly and then capitalizes on such selected proposals to distill the relation of each proposal in reference frame. Extensive experiments conducted on ImageNet VID dataset validate our proposal and analysis. More remarkably, we achieve to-date the best reported mAP of 84.7\%, after post-processing of linking and rescoring.

\paragraph{Acknowledgements} This work was supported in part by NSFC under contract No.~61836011, No.~61822208, and No. 61632019, and Youth Innovation Promotion Association CAS (No.~2018497).

{\small
\bibliographystyle{ieee_fullname}
\bibliography{egbib}

\begin{thebibliography}{10}\itemsep=-1pt

\bibitem{stsn}
Gedas Bertasius, Lorenzo Torresani, and Jianbo Shi.
\newblock Object detection in video with spatiotemporal sampling networks.
\newblock In {\em ECCV}, 2018.

\bibitem{cai2019exploring}
Qi Cai, Yingwei Pan, Chong-Wah Ngo, Xinmei Tian, Lingyu Duan, and Ting Yao.
\newblock Exploring object relation in mean teacher for cross-domain detection.
\newblock In {\em CVPR}, 2019.

\bibitem{Cai_2018_CVPR}
Zhaowei Cai and Nuno Vasconcelos.
\newblock Cascade r-cnn: Delving into high quality object detection.
\newblock In {\em CVPR}, 2018.

\bibitem{chen2014semantic}
Liang-Chieh Chen, George Papandreou, Iasonas Kokkinos, Kevin Murphy, and Alan~L
  Yuille.
\newblock Semantic image segmentation with deep convolutional nets and fully
  connected crfs.
\newblock {\em ICLR}, 2015.

\bibitem{dai2016r}
Jifeng Dai, Yi Li, Kaiming He, and Jian Sun.
\newblock R-fcn: Object detection via region-based fully convolutional
  networks.
\newblock In {\em NIPS}, 2016.

\bibitem{dai2017deformable}
Jifeng Dai, Haozhi Qi, Yuwen Xiong, Yi Li, Guodong Zhang, Han Hu, and Yichen
  Wei.
\newblock Deformable convolutional networks.
\newblock In {\em ICCV}, 2017.

\bibitem{dosovitskiy2015flownet}
Alexey Dosovitskiy, Philipp Fischer, Eddy Ilg, Philip Hausser, Caner Hazirbas,
  Vladimir Golkov, Patrick Van Der~Smagt, Daniel Cremers, and Thomas Brox.
\newblock Flownet: Learning optical flow with convolutional networks.
\newblock In {\em CVPR}, 2015.

\bibitem{feichtenhofer2017detect}
Christoph Feichtenhofer, Axel Pinz, and Andrew Zisserman.
\newblock Detect to track and track to detect.
\newblock In {\em ICCV}, 2017.

\bibitem{gehring2017convolutional}
Jonas Gehring, Michael Auli, David Grangier, Denis Yarats, and Yann~N Dauphin.
\newblock Convolutional sequence to sequence learning.
\newblock In {\em ICML}, 2017.

\bibitem{girshick2015fast}
Ross Girshick.
\newblock Fast r-cnn.
\newblock In {\em ICCV}, 2015.

\bibitem{girshick2014rich}
Ross Girshick, Jeff Donahue, Trevor Darrell, and Jitendra Malik.
\newblock Rich feature hierarchies for accurate object detection and semantic
  segmentation.
\newblock In {\em CVPR}, 2014.

\bibitem{gkioxari2015finding}
Georgia Gkioxari and Jitendra Malik.
\newblock Finding action tubes.
\newblock In {\em CVPR}, 2015.

\bibitem{han2016seq}
Wei Han, Pooya Khorrami, Tom~Le Paine, Prajit Ramachandran, Mohammad
  Babaeizadeh, Honghui Shi, Jianan Li, Shuicheng Yan, and Thomas~S Huang.
\newblock Seq-nms for video object detection.
\newblock {\em arXiv:1602.08465}, 2016.

\bibitem{he2017mask}
Kaiming He, Georgia Gkioxari, Piotr Doll{\'a}r, and Ross Girshick.
\newblock Mask r-cnn.
\newblock In {\em ICCV}, 2017.

\bibitem{he2015spatial}
Kaiming He, Xiangyu Zhang, Shaoqing Ren, and Jian Sun.
\newblock Spatial pyramid pooling in deep convolutional networks for visual
  recognition.
\newblock {\em IEEE Trans. on PAMI}, 2015.

\bibitem{he2016resnet}
Kaiming He, Xiangyu Zhang, Shaoqing Ren, and Jian Sun.
\newblock Deep residual learning for image recognition.
\newblock In {\em CVPR}, 2016.

\bibitem{hu2018relation}
Han Hu, Jiayuan Gu, Zheng Zhang, Jifeng Dai, and Yichen Wei.
\newblock Relation networks for object detection.
\newblock In {\em CVPR}, 2018.

\bibitem{hu2018squeeze}
Jie Hu, Li Shen, and Gang Sun.
\newblock Squeeze-and-excitation networks.
\newblock In {\em CVPR}, 2018.

\bibitem{huang2017densely}
Gao Huang, Zhuang Liu, Laurens Van Der~Maaten, and Kilian~Q Weinberger.
\newblock Densely connected convolutional networks.
\newblock In {\em CVPR}, 2017.

\bibitem{kang2017object}
Kai Kang, Hongsheng Li, Tong Xiao, Wanli Ouyang, Junjie Yan, Xihui Liu, and
  Xiaogang Wang.
\newblock Object detection in videos with tubelet proposal networks.
\newblock In {\em CVPR}, 2017.

\bibitem{kang2017t}
Kai Kang, Hongsheng Li, Junjie Yan, Xingyu Zeng, Bin Yang, Tong Xiao, Cong
  Zhang, Zhe Wang, Ruohui Wang, Xiaogang Wang, et~al.
\newblock T-cnn: Tubelets with convolutional neural networks for object
  detection from videos.
\newblock {\em IEEE Trans. on CSVT}, 2017.

\bibitem{krizhevsky2012imagenet}
Alex Krizhevsky, Ilya Sutskever, and Geoffrey~E Hinton.
\newblock Imagenet classification with deep convolutional neural networks.
\newblock In {\em NIPS}, 2012.

\bibitem{law2018cornernet}
Hei Law and Jia Deng.
\newblock Cornernet: Detecting objects as paired keypoints.
\newblock In {\em ECCV}, 2018.

\bibitem{li2018recurrent}
Dong Li, Zhaofan Qiu, Qi Dai, Ting Yao, and Tao Mei.
\newblock Recurrent tubelet proposal and recognition networks for action
  detection.
\newblock In {\em ECCV}, 2018.

\bibitem{li2018unified}
Dong Li, Ting Yao, Ling-Yu Duan, Tao Mei, and Yong Rui.
\newblock Unified spatio-temporal attention networks for action recognition in
  videos.
\newblock {\em IEEE Transactions on Multimedia}, 2018.

\bibitem{lin2017feature}
Tsung-Yi Lin, Piotr Doll{\'a}r, Ross~B Girshick, Kaiming He, Bharath Hariharan,
  and Serge~J Belongie.
\newblock Feature pyramid networks for object detection.
\newblock In {\em CVPR}, 2017.

\bibitem{Lin_2017_ICCV}
Tsung-Yi Lin, Priya Goyal, Ross Girshick, Kaiming He, and Piotr Dollar.
\newblock Focal loss for dense object detection.
\newblock In {\em ICCV}, 2017.

\bibitem{lin2014microsoft}
Tsung-Yi Lin, Michael Maire, Serge Belongie, James Hays, Pietro Perona, Deva
  Ramanan, Piotr Doll{\'a}r, and C~Lawrence Zitnick.
\newblock Microsoft {COCO}: Common objects in context.
\newblock In {\em ECCV}, 2014.

\bibitem{Liu_2018_CVPR}
Mason Liu and Menglong Zhu.
\newblock Mobile video object detection with temporally-aware feature maps.
\newblock In {\em CVPR}, 2018.

\bibitem{liu2016ssd}
Wei Liu, Dragomir Anguelov, Dumitru Erhan, Christian Szegedy, Scott Reed,
  Cheng-Yang Fu, and Alexander~C Berg.
\newblock Ssd: Single shot multibox detector.
\newblock In {\em ECCV}, 2016.

\bibitem{long2015fully}
Jonathan Long, Evan Shelhamer, and Trevor Darrell.
\newblock Fully convolutional networks for semantic segmentation.
\newblock In {\em CVPR}, 2015.

\bibitem{mallat1999wavelet}
St{\'e}phane Mallat.
\newblock {\em A wavelet tour of signal processing}.
\newblock 1999.

\bibitem{ouyang2014deepid}
Wanli Ouyang, Ping Luo, Xingyu Zeng, Shi Qiu, Yonglong Tian, Hongsheng Li, Shuo
  Yang, Zhe Wang, Yuanjun Xiong, Chen Qian, et~al.
\newblock Deepid-net: multi-stage and deformable deep convolutional neural
  networks for object detection.
\newblock {\em arXiv:1409.3505}, 2014.

\bibitem{pan2016learning}
Yingwei Pan, Yehao Li, Ting Yao, Tao Mei, Houqiang Li, and Yong Rui.
\newblock Learning deep intrinsic video representation by exploring temporal
  coherence and graph structure.
\newblock In {\em IJCAI}, 2016.

\bibitem{pan2016jointly}
Yingwei Pan, Tao Mei, Ting Yao, Houqiang Li, and Yong Rui.
\newblock Jointly modeling embedding and translation to bridge video and
  language.
\newblock In {\em CVPR}, 2016.

\bibitem{paszke2017automatic}
Adam Paszke, Sam Gross, Soumith Chintala, Gregory Chanan, Edward Yang, Zachary
  DeVito, Zeming Lin, Alban Desmaison, Luca Antiga, and Adam Lerer.
\newblock Automatic differentiation in pytorch.
\newblock In {\em Workshop on Machine Learning Systems, NIPS}, 2017.

\bibitem{qiu2017learning}
Zhaofan Qiu, Ting Yao, and Tao Mei.
\newblock Learning spatio-temporal representation with pseudo-3d residual
  networks.
\newblock In {\em ICCV}, 2017.

\bibitem{redmon2016you}
Joseph Redmon, Santosh Divvala, Ross Girshick, and Ali Farhadi.
\newblock You only look once: Unified, real-time object detection.
\newblock In {\em CVPR}, 2016.

\bibitem{ren2015faster}
Shaoqing Ren, Kaiming He, Ross Girshick, and Jian Sun.
\newblock Faster r-cnn: Towards real-time object detection with region proposal
  networks.
\newblock In {\em NIPS}, 2015.

\bibitem{ILSVRC15}
Olga Russakovsky, Jia Deng, Hao Su, Jonathan Krause, Sanjeev Satheesh, Sean Ma,
  Zhiheng Huang, Andrej Karpathy, Aditya Khosla, Michael Bernstein,
  Alexander~C. Berg, and Li Fei-Fei.
\newblock {ImageNet Large Scale Visual Recognition Challenge}.
\newblock {\em IJCV}, 2015.

\bibitem{sermanet2014overfeat}
Pierre Sermanet, David Eigen, Xiang Zhang, Micha{\"e}l Mathieu, Robert Fergus,
  and Yann Lecun.
\newblock Overfeat: Integrated recognition, localization and detection using
  convolutional networks.
\newblock In {\em ICLR}, 2014.

\bibitem{simonyan2014two}
Karen Simonyan and Andrew Zisserman.
\newblock Two-stream convolutional networks for action recognition in videos.
\newblock In {\em NIPS}, 2014.

\bibitem{Simonyan:ICLR15}
Karen Simonyan and Andrew Zisserman.
\newblock Very deep convolutional networks for large-scale image recognition.
\newblock In {\em ICLR}, 2015.

\bibitem{Szegedy:CVPR15}
Christian Szegedy, Wei Liu, Yangqing Jia, Pierre Sermanet, Scott~E. Reed,
  Dragomir Anguelov, Dumitru Erhan, Vincent Vanhoucke, and Andrew Rabinovich.
\newblock Going deeper with convolutions.
\newblock In {\em CVPR}, 2015.

\bibitem{vaswani2017attention}
Ashish Vaswani, Noam Shazeer, Niki Parmar, Jakob Uszkoreit, Llion Jones,
  Aidan~N Gomez, {\L}ukasz Kaiser, and Illia Polosukhin.
\newblock Attention is all you need.
\newblock In {\em NIPS}, 2017.

\bibitem{wang2018manet}
Shiyao Wang, Yucong Zhou, Junjie Yan, and Zhidong Deng.
\newblock Fully motion-aware network for video object detection.
\newblock In {\em ECCV}, 2018.

\bibitem{wang2018non}
Xiaolong Wang, Ross Girshick, Abhinav Gupta, and Kaiming He.
\newblock Non-local neural networks.
\newblock In {\em CVPR}, 2018.

\bibitem{wang2018videos}
Xiaolong Wang and Abhinav Gupta.
\newblock Videos as space-time region graphs.
\newblock In {\em ECCV}, 2018.

\bibitem{xiao2018matchtrans}
Fanyi Xiao and Yong Jae~Lee.
\newblock Video object detection with an aligned spatial-temporal memory.
\newblock In {\em ECCV}, 2018.

\bibitem{xie2017aggregated}
Saining Xie, Ross Girshick, Piotr Doll{\'a}r, Zhuowen Tu, and Kaiming He.
\newblock Aggregated residual transformations for deep neural networks.
\newblock In {\em CVPR}, 2017.

\bibitem{yang2016craft}
Bin Yang, Junjie Yan, Zhen Lei, and Stan~Z Li.
\newblock Craft objects from images.
\newblock In {\em CVPR}, 2016.

\bibitem{yao2018exploring}
Ting Yao, Yingwei Pan, Yehao Li, and Tao Mei.
\newblock Exploring visual relationship for image captioning.
\newblock In {\em ECCV}, 2018.

\bibitem{zhu2018towards}
Xizhou Zhu, Jifeng Dai, Lu Yuan, and Yichen Wei.
\newblock Towards high performance video object detection.
\newblock In {\em CVPR}, 2018.

\bibitem{zhu2017fgfa}
Xizhou Zhu, Yujie Wang, Jifeng Dai, Lu Yuan, and Yichen Wei.
\newblock Flow-guided feature aggregation for video object detection.
\newblock In {\em CVPR}, 2017.

\bibitem{zhu2017dff}
Xizhou Zhu, Yuwen Xiong, Jifeng Dai, Lu Yuan, and Yichen Wei.
\newblock Deep feature flow for video recognition.
\newblock In {\em CVPR}, 2017.

\end{thebibliography}
}

\end{document}